\documentclass[11pt,a4paper]{article}

\usepackage[T1]{fontenc}
\usepackage[utf8]{inputenc}
\usepackage{lmodern}
\usepackage{microtype}
\usepackage[a4paper,margin=2.6cm]{geometry}
\usepackage{setspace}
\usepackage{parskip}
\usepackage{enumitem}
\usepackage{booktabs}
\usepackage{array}
\usepackage{siunitx}
\usepackage{graphicx}
\usepackage{subcaption}
\usepackage{float}
\usepackage{placeins}
\usepackage{amsmath,amssymb}
\usepackage{xcolor}
\usepackage{csvsimple}
\usepackage{csquotes}
\usepackage[toc,page]{appendix}
\usepackage[hidelinks]{hyperref}
\usepackage[nameinlink,noabbrev]{cleveref}
\usepackage[
  backend=biber,
  style=authoryear,
  sorting=nyt,
  maxcitenames=2,
  maxbibnames=99
]{biblatex}

\graphicspath{{results_figures/}}

\crefname{appendix}{Appendix}{Appendices}
\Crefname{appendix}{Appendix}{Appendices}

\setlist[itemize]{topsep=0.4em,itemsep=0.25em}
\setlist[enumerate]{topsep=0.4em,itemsep=0.25em}

\newcommand{\RRF}{\mathrm{RRF}}
\newcommand{\MRR}{\mathrm{MRR}}
\newcommand{\RR}{\mathrm{RR}}
\newcommand{\dataset}[1]{\texttt{\detokenize{#1}}}

\newcommand{\repositoryurl}{\url{https://github.com/pereirabarataap/dutch-mteb-hybrid-rrf}}

\newcommand{\resultgraphic}[2][\linewidth]{%
  \IfFileExists{results_figures/#2}{%
    \includegraphics[width=#1]{#2}%
  }{%
    \fbox{\parbox[c][4.5cm][c]{0.92\linewidth}{\centering
      Missing figure: \path{results_figures/#2}}}%
  }%
}

\title{%
  \textbf{Do Static Embeddings Add Value\\to Hybrid Dutch Retrieval?}\\[0.5em]
  \large Cross-Validated Weighted RRF with Paired Inference and Cross-Domain Transfer
}
\author{António Pereira Barata}
\date{August 2026}

\begin{document}

\maketitle

\begin{abstract}
Embedding benchmarks measure standalone model quality, but they do not establish whether a low-cost retriever contributes complementary ranking information once lexical and transformer-based retrieval are already combined. We present a controlled evaluation of this question across Dutch retrieval tasks from the Massive Text Embedding Benchmark for Dutch (MTEB-NL). Weighted reciprocal rank fusion (RRF) combines Best Matching 25 (BM25), Qwen/Qwen3-Embedding-0.6B (Qwen), and two multilingual static embedding models. Five datasets comprising 14,500 queries and 786,573 documents are scored exhaustively, and fusion weights are searched on a simplex in increments of 0.1. Ten-fold query-level cross-validation selects weights on nine folds and evaluates them on the held-out fold; paired bootstrap confidence intervals and sign-randomisation tests quantify the resulting differences. Fusion improves over the training-selected individual retriever by 0.061 mean reciprocal rank (MRR) on Dutch News, 0.029 on VABB, 0.004 on WebFAQ NL, and 0.025 on Wikipedia NL, while matching BM25 on Open Tender. All four positive differences remain distinguishable from zero after Holm correction. No unrestricted fold assigns positive weight to either static retriever: all 50 selections lie on the BM25--Qwen edge, and forcing a static contribution reduces effectiveness. Leave-one-dataset-out selection chooses equal BM25--Qwen weighting in every iteration and outperforms the cross-domain-selected individual retriever on every held-out task. The results support a two-retriever lexical–transformer architecture as a robust tested default across the evaluated Dutch tasks and show that standalone benchmark performance is insufficient to establish marginal value in hybrid retrieval.
\end{abstract}


\clearpage
\tableofcontents
\clearpage

\section{Introduction}
\label{sec:introduction}
Modern retrieval systems increasingly combine exact lexical matching with semantic representations. Lexical methods such as BM25 remain strong when relevance depends on terminology, named entities, or rare tokens, whereas dense encoders can retrieve semantically related documents even when query and document wording differs. These strengths are complementary, but the effectiveness of their combination depends on the data domain, query formulation, and ranking behaviour of the component systems.

Embedding benchmarks such as MTEB and MTEB-NL primarily compare models as standalone encoders across fixed tasks \parencite{muennighoff2022mteb,banar2025mtebnl}. Such comparisons are necessary, but they do not answer a different systems question: whether an additional retriever contributes ranking information that is not already supplied by a lexical--transformer hybrid. A weaker standalone model may still improve a fusion if its errors are complementary, while a competitive standalone model may add no marginal value. This distinction is especially relevant for modern static embeddings, which trade contextual expressiveness for substantially lower encoding cost.

This paper studies that marginal-value question for Dutch retrieval. Weighted RRF is applied to BM25, Qwen/Qwen3-Embedding-0.6B, and two multilingual static embedding models over five MTEB-NL retrieval datasets. This is a controlled evaluation of whether modern multilingual static embeddings improve a BM25--transformer fusion across Dutch MTEB retrieval tasks, while also estimating both within-dataset selection performance and transfer to an unseen dataset. The objective is not to propose a new fusion rule, but to establish which retrieval components are empirically justified under held-out evaluation.

The evaluation is deliberately exhaustive. Every query is scored against every document in its dataset, so fusion is computed from complete rankings rather than truncated candidate lists. This removes top-$k$ approximation as a confounder and permits exact calculation of the relevant-document rank for every evaluated weight configuration.

This paper makes four contributions:
\begin{itemize}
  \item a controlled test of whether two modern multilingual static retrievers add complementary effectiveness beyond BM25--Qwen fusion on five Dutch MTEB retrieval tasks;
  \item exhaustive full-corpus scoring and complete simplex searches over weighted lexical, transformer, and static rankings;
  \item training-only weight selection with out-of-fold query scores, paired uncertainty estimates, multiplicity-corrected tests, and leave-one-dataset-out transfer; and
  \item evidence that the selected optimum always lies on the BM25--Qwen edge, together with a transferable equal-weight BM25--Qwen default for settings without in-domain tuning data.
\end{itemize}

The remainder of this paper is structured as follows. \Cref{sec:background} positions the study and introduces the retrieval methods, fusion procedure, and benchmark; \cref{sec:research-problem-questions} presents the research problem and questions. \Cref{sec:experimental-method} describes the experimental design, and \cref{sec:results,sec:discussion} report and interpret the findings. \Cref{sec:limitations,sec:recommended-extensions} discuss limitations and extensions before \cref{sec:conclusion} concludes. \Cref{sec:data-code-availability} identifies the generated artefacts, while \cref{app:simplex-heatmaps,app:reproducibility} provide the complete simplex visualisations and reproducibility details.

\section{Background and Study Positioning}
\label{sec:background}
This section introduces the retrieval concepts and prior evaluation setting needed to interpret the study. \Cref{subsec:background-retrieval} describes lexical, contextual, and static retrieval signals. \Cref{subsec:background-complementarity} distinguishes standalone benchmarking from testing marginal value in a hybrid system. \Cref{subsec:background-rrf} reviews rank-based fusion, and \cref{subsec:background-benchmark} describes the Dutch benchmark setting and effectiveness metric.

\subsection{Complementary retrieval signals}
\label{subsec:background-retrieval}
Retrieval systems must balance exact matching against semantic generalisation. Lexical methods such as BM25 rank documents from query-term occurrences and remain effective when relevance depends on terminology, identifiers, named entities, or rare words \parencite{robertson2009bm25}. Dense retrieval instead represents queries and documents in a shared vector space, allowing semantically related documents to be retrieved despite limited vocabulary overlap. Transformer encoders such as Qwen provide contextual representations, whereas static embedding models offer substantially cheaper but less expressive semantic representations \parencite{zhang2025qwen3embedding,tulkens2024model2vec,minishlab2025potion,aarsen2024staticsimilarity}.

These approaches fail in different ways. BM25 may miss paraphrases, while dense models may underweight exact lexical evidence or rare domain-specific expressions. A static model may be weaker individually yet still contribute if its representation space produces different ranking errors. The experiments consequently compare BM25, Qwen, minishlab/potion-multilingual-128M (Potion), and tomaarsen/static-similarity-mrl-multilingual-v1 (Static Similarity) both as standalone retrievers and as fusion components. Potion is distributed with broad MTEB evaluations, whereas the Static Similarity model card explicitly positions that model outside retrieval use cases; it is included as a second modern multilingual static representation rather than as a retrieval-specialised baseline \parencite{minishlab2025potion,aarsen2024staticsimilarity}. Their implementations and preprocessing choices are described in \cref{subsec:method-retrievers}.

\subsection{Standalone benchmarks and marginal fusion value}
\label{subsec:background-complementarity}
MTEB-style leaderboards estimate how well individual embedding models perform across tasks \parencite{muennighoff2022mteb,banar2025mtebnl}. They do not determine whether two rankings contain complementary information when combined with a lexical retriever. This marginal contribution must be measured directly: standalone quality and incremental fusion value are related but distinct properties.

Hybrid lexical--dense retrieval and RRF are established techniques
\parencite{kuzi2020hybrid,cormack2009rrf}. Prior work has evaluated
lexical, dense, and reranking approaches for Dutch retrieval
\parencite{lotfi2025beirnl}. A public Danish Model2Vec evaluation has
also reported BM25--static-embedding fusion using RRF
\parencite{andersborges2025model2vecdkstem}. The contribution of the
present work therefore does not lie in introducing BM25--dense fusion.
The empirical question is narrower: whether low-cost multilingual static
rankings improve an already heterogeneous BM25--transformer system on
Dutch tasks, whether any selected contribution survives held-out
evaluation, and whether the resulting weights transfer across datasets.

\subsection{Combining heterogeneous retrieval outputs}
\label{subsec:background-rrf}
Lexical and semantic retrievers produce scores with different scales and distributions, so their outputs cannot generally be added without calibration. Possible combination strategies include score normalisation followed by linear interpolation, supervised learning-to-rank, reranking a shared candidate set, and rank-based fusion.

Score-level methods retain information about score magnitude but depend on a suitable normalisation procedure. Supervised fusion can learn more flexible combinations, but requires training data and introduces an additional fitted model.

Rank fusion instead combines the relative ordering produced by each retriever. RRF is widely used for hybrid retrieval because it is simple, robust to incompatible score scales, and can be applied without additional training \parencite{cormack2009rrf}. These properties make it suitable for the present objective: determining whether the component rankings contain complementary information rather than learning a new retrieval model.

For document $d$ and query $q$, the weighted three-retriever score is
\begin{equation}
\RRF(d\mid q) = \frac{\alpha}{k+r_{\text{BM25}}(d\mid q)}
+ \frac{\beta}{k+r_{\text{Qwen}}(d\mid q)}
+ \frac{\gamma}{k+r_{\text{static}}(d\mid q)},
  \label{eq:rrf}
\end{equation}
where $r(d\mid q)$ denotes the one-based rank and
\begin{equation}
  \alpha+\beta+\gamma=1,
  \qquad
  \alpha,\beta,\gamma\geq 0.
\end{equation}
The constant $k$ controls how strongly the highest ranks dominate the contribution of each system. The selected value and weight-search procedure are specified in \cref{subsec:method-scoring-fusion}.

\subsection{Evaluating retrieval effectiveness}
\label{subsec:background-benchmark}
Retrieval systems can be evaluated through online user experiments, manually constructed domain-specific test collections, or established benchmark suites. Online evaluation measures operational impact directly but requires a deployed system and sufficient interaction data. Bespoke collections permit close alignment with a target domain, but their construction is costly and their results may be difficult to compare with prior work. Standard benchmarks instead provide fixed corpora, queries, and relevance judgements, enabling reproducible comparison across retrieval methods.

MTEB provides a common evaluation framework for embedding and retrieval models across tasks and languages \parencite{muennighoff2022mteb}. MTEB-NL extends this framework with Dutch datasets covering different domains, corpus sizes, and query formulations \parencite{banar2025mtebnl}. The public benchmark data make the present comparison reproducible; the newly evaluated objects are the exhaustive component rankings, their weighted combinations, and the held-out selection and transfer outcomes.

Each selected task contains exactly one relevant document per query. Under this condition, average precision (AP) is numerically identical to reciprocal rank (RR), while binary normalised discounted cumulative gain (nDCG) is likewise determined solely by the relevant document's rank, albeit using a different discount function. We therefore use mean reciprocal rank (MRR) as the primary performance measure rather than reporting multiple metrics. The selected datasets, relevance mappings, and validation procedures are detailed in \cref{subsec:method-datasets,subsec:method-evaluation}.

\section{Research Problem and Questions}
\label{sec:research-problem-questions}
This section formulates the central problem of this work (\Cref{subsec:problem-statement}) and translates it into a set of empirically answerable questions (\Cref{subsec:research-questions}).

\subsection{Problem Statement}
\label{subsec:problem-statement}

Dutch retrieval tasks differ in terminology, query style, corpus size, and dependence on lexical overlap or semantic similarity. Standalone benchmark performance does not establish whether a retriever adds information beyond the rankings already supplied by other components. Additional retrievers also increase storage, inference, and operational complexity, so their inclusion should be supported by held-out marginal effectiveness rather than an in-sample optimum.

\begin{quote}
  \textbf{Problem statement.} Which combination of BM25, transformer-based semantic retrieval, and low-cost static embeddings provides the strongest expected retrieval effectiveness on unseen Dutch queries, and which conclusions remain valid when the selected configuration is transferred to an unseen dataset?
\end{quote}

The analysis therefore separates three quantities: the best configuration observed on all available queries, the expected performance of a configuration-selection procedure on unseen queries from the same dataset, and the performance of a configuration selected without using the target dataset. It also distinguishes standalone retriever quality from complementary value inside a fused system.

\subsection{Research Questions}
\label{subsec:research-questions}

The problem statement is made empirically tractable through the following research questions:

\begin{enumerate}[label=\textbf{RQ\arabic*.},leftmargin=*]
  \item Does cross-validated weighted RRF outperform the individual retriever selected from the same training queries?
  \item Does either of the static embedding models add complementary effectiveness beyond BM25--Qwen fusion?
  \item How stable are the selected fusion weights, and how large is the gap between apparent full-data performance and out-of-fold performance?
  \item Does a configuration selected on other datasets transfer to an unseen Dutch retrieval task?
  \item How do the evaluated fixed and cross-validated approaches rank by macro-average MRR across datasets?
\end{enumerate}

\section{Experimental Method}
\label{sec:experimental-method}
This section describes the complete evaluation procedure. \Cref{subsec:method-datasets,subsec:method-retrievers} specify the datasets, preparation steps, and retrieval systems; \cref{subsec:method-scoring-fusion,subsec:method-evaluation} define exhaustive scoring, fusion, and MRR. \Cref{subsec:method-cross-validation,subsec:method-selection-stability,subsec:method-paired-inference} describe within-dataset CV estimation and inference, while \cref{subsec:method-cross-domain,subsec:method-simplex-surfaces} cover cross-domain transfer and descriptive simplex analysis.

\subsection{Datasets and data preparation}
\label{subsec:method-datasets}
The first evaluation split exposed by each MTEB task is loaded. Document titles are prepended to bodies when available so that every retriever receives the same text. Queries and documents are converted to strings, deduplicated, and sorted by dataset and identifier. The fusion stage verifies that each query has exactly one relevant document.

\begin{table}[H]
  \centering
  \caption{Retrieval datasets and evaluated split sizes.}
  \label{tab:datasets}
  \small
  \begin{tabular}{@{}llrrl@{}}
    \toprule
    MTEB task & Internal name & Queries & Documents & Query source \\
    \midrule
    DutchNewsArticlesRetrieval & \dataset{dutch_news} & 1,000 & 255,524 & Human-written \\
    OpenTenderRetrieval & \dataset{open_tender} & 1,000 & 137,633 & Human-written \\
    VABBRetrieval & \dataset{vabb} & 1,000 & 9,254 & Human-written \\
    WebFAQRetrieval (nld) & \dataset{webfaq_nl} & 10,000 & 370,662 & Human-written \\
    WikipediaRetrievalMultilingual (nl) & \dataset{wikipedia_nl} & 1,500 & 13,500 & LLM-generated \\
    \midrule
    \textbf{Total} & & \textbf{14,500} & \textbf{786,573} & \\
    \bottomrule
  \end{tabular}
\end{table}

WebFAQ NL contributes 10,000 of the 14,500 queries. Cross-dataset selection and method ranking therefore use dataset-level macro averages so that WebFAQ NL does not dominate solely through its query count.

\subsection{Retrieval systems}
\label{subsec:method-retrievers}
\begin{table}[H]
  \centering
  \caption{Evaluated retrieval systems.}
  \label{tab:retrievers}
  \small
  \begin{tabular}{@{}p{0.29\linewidth}p{0.16\linewidth}p{0.11\linewidth}p{0.36\linewidth}@{}}
    \toprule
    Retriever & Family & Dimension & Configuration \\
    \midrule
    BM25 & Lexical & Sparse & Dutch stopwords and stemming \\
    Qwen & Transformer dense & 1024 & Query instruction for queries; no instruction for documents \\
    Potion & Static dense & 256 & Model2Vec encoding \\
    Static Similarity & Static dense & 1024 & Sentence Transformers static encoding \\
    \bottomrule
  \end{tabular}
\end{table}

All dense embeddings are stored as \texttt{float32} arrays and $L_2$-normalised. Dot products are therefore equivalent to cosine similarity. Alignment metadata records row counts, dimensions, data types, and hashes to guard against mismatches between embedding rows and identifiers. The recorded execution environment, software versions, and embedding-matrix dimensions are reported in Appendix~\ref{app:reproducibility}.

\subsection{Exhaustive scoring and fusion}
\label{subsec:method-scoring-fusion}
Every query is scored against every document. Dense scores are computed as batched matrix products, while BM25 requests a result count equal to the full corpus size. The experiment evaluates
\begin{equation}
  \sum_{j=1}^{5} |Q_j|\,|D_j| = 4{,}129{,}281{,}000
\end{equation}
query--document pairs per retriever, or 16,517,124,000 score rows across four retrievers.

The exhaustive score artifacts occupy approximately 103 gigabytes (GB) as reported by Windows. Exact logical and on-disk byte counts are given in Appendix~\ref{app:reproducibility}. 

Complete wall-clock measurements are available for scoring and fusion/ranking, which together account for 3:30:56 of measured execution time. Because corresponding measurements are unavailable for data preparation, embedding generation, index construction, and final statistical analysis, this value is a measured lower bound rather than an end-to-end runtime; the available breakdown is also reported in Appendix~\ref{app:reproducibility}.

Scores are ranked independently for each query and model using Structured Query Language (SQL) \texttt{RANK} semantics. Weighted RRF uses $k=60$ and non-negative weights sampled in increments of 0.1. The grid contains 66 triples per static-model panel. Configurations with $\gamma=0$ are duplicated across panels by construction; their query-level scores are verified to agree and are collapsed before selection.

\subsection{Effectiveness metric}
\label{subsec:method-evaluation}

For query $q$ with one relevant document at rank $r(q)$, reciprocal rank (RR) and mean reciprocal rank (MRR) are defined as
\begin{equation}
  \RR(q)=\frac{1}{r(q)},
  \qquad
  \MRR=\frac{1}{|Q|}\sum_{q\in Q}\RR(q).
  \label{eq:mrr}
\end{equation}
Because each query has exactly one relevant document, MRR is used as the sole effectiveness metric.

\subsection{Ten-fold estimation of the selection procedure}
\label{subsec:method-cross-validation}
Queries are randomly partitioned into ten deterministic, disjoint folds within each dataset. For fold $f$, every fusion configuration is averaged over the other nine folds, the highest-training-MRR configuration is selected, and that configuration is applied unchanged to fold $f$. After ten iterations, every query has exactly one out-of-fold RR value generated without using that query for weight selection.

The reported fusion estimate is the mean of the ten held-out fold means. Fixed retrievers require no configuration selection, so their pooled out-of-fold mean is equal to their ordinary all-query mean; they are nevertheless evaluated over the same folds to obtain a comparable fold-level spread. Error bars show one sample standard deviation across the ten test-fold means. This standard deviation describes variation between held-out subsets and is not presented as a standard error or CI.

The strongest individual baseline is also selected inside each training split from BM25, Qwen, Potion, and Static Similarity, then deployed on the same held-out queries as the selected fusion. This avoids choosing the baseline after observing the test fold.

\subsection{Weight stability and model-selection bias}
\label{subsec:method-selection-stability}
For each dataset, the analysis records the selected configuration in every fold, its frequency, and the mean and standard deviation of $\alpha$, $\beta$, and $\gamma$. Selecting the best simplex configuration and evaluating it on the same queries yields an apparent, or resubstitution, performance estimate and can introduce model-selection bias. The full-data optimum is therefore retained only as a descriptive diagnostic. The reported gap is
\begin{equation}
  \Delta_{\mathrm{selection}}
  = \MRR_{\mathrm{apparent\ full\text{-}data}}
  - \MRR_{\mathrm{10\text{-}fold\ CV}}.
\end{equation}
A positive value indicates that same-sample selection and evaluation overstate the held-out estimate. A value near zero means that the same configuration was sufficiently stable in this experiment; it does not remove the need for CV.

\subsection{Paired uncertainty and hypothesis tests}
\label{subsec:method-paired-inference}
Every out-of-fold query receives scores from both methods in a comparison, so inference is performed on paired RR differences. A percentile paired bootstrap CI estimates the plausible magnitude of the mean MRR difference. A two-sided sign-randomisation test evaluates the null hypothesis that the paired effects are exchangeable around zero.

The primary comparison asks whether CV-selected fusion improves over the individual retriever selected on the same training folds. Additional comparisons test whether allowing a static ranker changes the selected fusion and whether the best positive-static-weight subspace improves over the BM25--Qwen edge. Holm adjustment is applied across the five dataset-level tests within each comparison. The $p$-values address whether a difference is difficult to reconcile with a zero-effect null; the mean difference and CI determine its practical magnitude.

\subsection{Leave-one-dataset-out transfer}
\label{subsec:method-cross-domain}
Query-level CV estimates performance on new queries from a known dataset. It does not determine whether weights transfer to a new domain. Leave-one-dataset-out evaluation therefore selects one fusion configuration by macro-average MRR on four datasets and deploys it unchanged on the fifth. The individual-retriever baseline is selected by the same four-dataset macro average. Repeating this procedure for all five held-out datasets measures cross-domain transfer without using the target dataset for either selection decision.

\subsection{Descriptive simplex surfaces}
\label{subsec:method-simplex-surfaces}
The complete full-query simplex heatmaps are retained to explain the shape of the search space and the location of local and global optima. They are descriptive visualisations, not substitutes for the CV or leave-one-dataset-out estimates.

\section{Results}
\label{sec:results}
The central empirical result is that every selected optimum lies on the BM25--Qwen edge: neither static retriever receives positive weight in any of the 50 held-out selections. \Cref{subsec:results-within-dataset} first presents expected within-dataset performance, \cref{subsec:results-paired-gains} quantifies paired fusion gains, and \cref{subsec:results-stability-static} tests static-model contribution directly. \Cref{subsec:results-cross-domain,subsec:results-method-ordering} then report cross-domain transfer and the macro-average ordering of methods.

\subsection{Expected within-dataset performance}
\label{subsec:results-within-dataset}
The expected performance of the selection procedure is higher than the selected individual retriever on four datasets and equal on Open Tender. Because the same configuration is selected in all ten folds of every dataset, the CV means equal the corresponding full-query optima to the displayed precision. This agreement is an empirical result of stable selection, not a justification for evaluating tuned configurations in sample.

\begin{figure}[H]
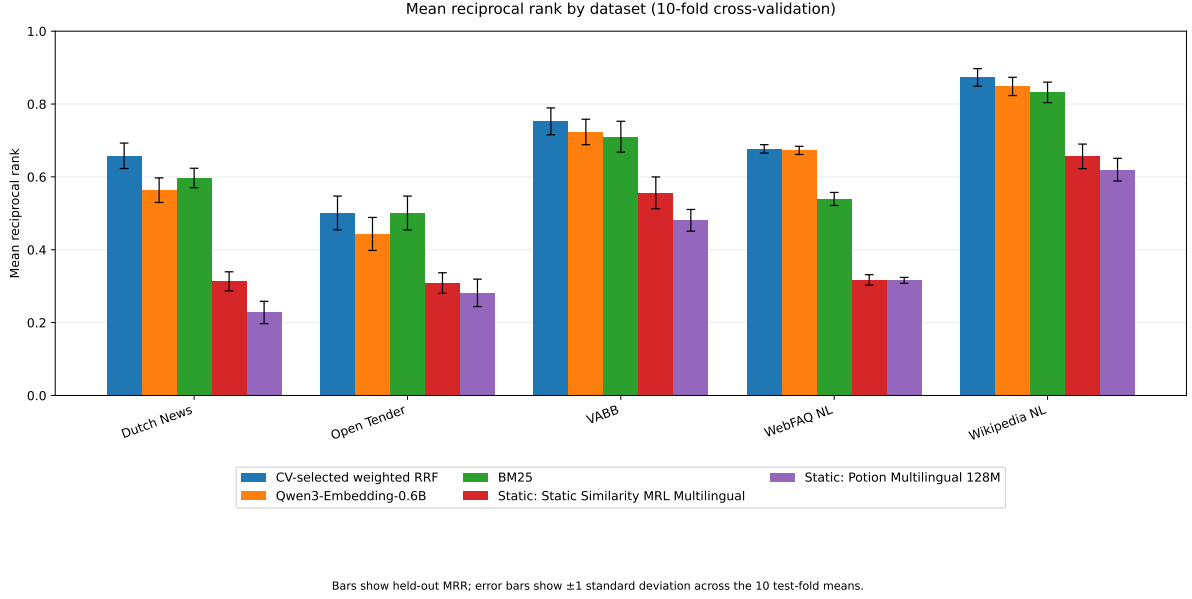

  \centering
  \resultgraphic{dataset_mrr_10fold_cv_with_fold_sd.pdf}
  \caption{Held-out MRR by dataset and retrieval method. Fusion weights are selected inside each training split. Error bars show $\pm 1$ sample standard deviation across the ten test-fold means.}
  \label{fig:dataset-bars}
\end{figure}

\begin{table}[H]
  \centering
  \caption{Mean MRR with sample standard deviation across the ten test-fold means in parentheses. Fixed-method means equal their all-query means; fusion is generated entirely out of fold after training-only weight selection.}
  \label{tab:all-methods}
  \scriptsize
  \begin{tabular}{@{}lccccc@{}}
    \toprule
    Dataset & BM25 & Qwen & Potion & Static Similarity & CV-selected fusion \\
    \midrule
    Dutch News   & 0.597 (0.027) & 0.563 (0.034) & 0.228 (0.031) & 0.313 (0.026) & 0.658 (0.035) \\
    Open Tender  & 0.501 (0.046) & 0.443 (0.045) & 0.282 (0.038) & 0.309 (0.028) & 0.501 (0.046) \\
    VABB         & 0.710 (0.042) & 0.723 (0.035) & 0.481 (0.030) & 0.556 (0.044) & 0.752 (0.037) \\
    WebFAQ NL    & 0.539 (0.018) & 0.673 (0.011) & 0.316 (0.008) & 0.317 (0.014) & 0.677 (0.012) \\
    Wikipedia NL & 0.832 (0.028) & 0.848 (0.025) & 0.620 (0.031) & 0.656 (0.034) & 0.873 (0.024) \\
    \bottomrule
  \end{tabular}
\end{table}

\begin{table}[H]
  \centering
  \caption{Primary 10-fold comparison. The baseline retriever is selected on each training split.}
  \label{tab:cv-primary}
  \small
  \begin{tabular}{@{}lrlrr@{}}
    \toprule
    Dataset & Fusion MRR & Selected individual & Individual MRR & Difference \\
    \midrule
    Dutch News & 0.658 & BM25 & 0.597 & 0.061 \\
    Open Tender & 0.501 & BM25 & 0.501 & 0.000 \\
    VABB & 0.752 & Qwen & 0.723 & 0.029 \\
    WebFAQ NL & 0.677 & Qwen & 0.673 & 0.004 \\
    Wikipedia NL & 0.873 & Qwen & 0.848 & 0.025 \\
    \bottomrule
  \end{tabular}
\end{table}

\subsection{Paired evidence for the fusion gains}
\label{subsec:results-paired-gains}
The positive gains on Dutch News, VABB, WebFAQ NL, and Wikipedia NL remain distinguishable from zero after multiplicity correction. The WebFAQ NL CI is entirely positive, but its absolute gain is only 0.004 MRR. Statistical evidence therefore supports a non-zero effect while the effect-size estimate warns against overstating its operational importance.

\begin{table}[H]
  \centering
  \caption{Paired out-of-fold fusion improvements over the training-selected individual retriever. 95\% CIs are paired bootstrap intervals; $p_{\mathrm{Holm}}$ is the sign-randomisation $p$-value adjusted across the five datasets.}
  \label{tab:paired-primary}
  \small
  \begin{tabular}{@{}lrrrr@{}}
    \toprule
    Dataset & $\Delta$MRR & CI lower & CI upper & $p_{\mathrm{Holm}}$ \\
    \midrule
    Dutch News & 0.0610 & 0.0431 & 0.0792 & 0.0005 \\
    Open Tender & 0.0000 & 0.0000 & 0.0000 & 1.0000 \\
    VABB & 0.0292 & 0.0139 & 0.0445 & 0.0012 \\
    WebFAQ NL & 0.0041 & 0.0011 & 0.0071 & 0.0138 \\
    Wikipedia NL & 0.0246 & 0.0136 & 0.0354 & 0.0005 \\
    \bottomrule
  \end{tabular}
\end{table}

\subsection{Weight stability and static-model contribution}
\label{subsec:results-stability-static}
The unrestricted selector chooses one configuration in all ten folds of each dataset: $(0.5,0.5,0.0)$ for Dutch News, VABB, and Wikipedia NL; $(1.0,0.0,0.0)$ for Open Tender; and $(0.1,0.9,0.0)$ for WebFAQ NL. The number of distinct selected configurations is one for every dataset, the modal selection rate is 1.0, and the apparent-minus-CV gap is 0.000 to the reported precision.

No unrestricted fold assigns positive weight to a static retriever: all 50 selections have $\gamma=0$. Forcing a static-model contribution therefore reduces performance relative to the BM25--Qwen edge. The MRR decrease is 0.015--0.032 for Potion and 0.010--0.036 for Static Similarity. All losses remain significant after Holm correction except Static Similarity on VABB ($\Delta\MRR=-0.0099$, 95\% CI $[-0.0202,0.0003]$, $p_{\mathrm{Holm}}=0.0536$).

\subsection{Cross-domain transfer}
\label{subsec:results-cross-domain}
Leave-one-dataset-out selection chooses $(\alpha,\beta,\gamma)=(0.5,0.5,0.0)$ in every iteration. The individual baseline selected on the four training datasets is Qwen for four held-out tasks and BM25 for WebFAQ NL.

\begin{table}[H]
  \centering
  \caption{Leave-one-dataset-out transfer. Both the fusion configuration and individual baseline are selected without the held-out dataset. Transfer cost is the within-dataset CV fusion MRR minus transferred fusion MRR.}
  \label{tab:lodo}
  \small
  \begin{tabular}{@{}lrrrr@{}}
    \toprule
    Held-out dataset & Fusion MRR & Individual MRR & Difference & Transfer cost \\
    \midrule
    Dutch News & 0.658 & 0.563 & 0.094 & 0.000 \\
    Open Tender & 0.480 & 0.443 & 0.037 & 0.021 \\
    VABB & 0.752 & 0.723 & 0.029 & 0.000 \\
    WebFAQ NL & 0.651 & 0.539 & 0.112 & 0.026 \\
    Wikipedia NL & 0.873 & 0.848 & 0.025 & 0.000 \\
    \bottomrule
  \end{tabular}
\end{table}

Equal BM25--Qwen fusion transfers better than the individual retriever chosen from the four training datasets on all five held-out tasks. Its macro-average held-out MRR is 0.683, compared with 0.624 for the cross-domain-selected individual baseline, a mean difference of 0.059. It also reproduces the within-dataset selected result on Dutch News, VABB, and Wikipedia NL. The transfer costs on Open Tender and WebFAQ NL show the value of in-domain tuning when representative queries are available: Open Tender prefers pure BM25, whereas WebFAQ NL prefers a Qwen-heavy mixture.

\subsection{Cross-dataset method ordering}
\label{subsec:results-method-ordering}
The macro-average ordering of the main bars is CV-selected fusion first (0.692), followed by Qwen (0.650), BM25 (0.636), Static Similarity (0.430), and Potion (0.385). The fusion entry averages held-out predictions from a training-only selection procedure. It still represents dataset-adaptive weights rather than one fixed universal vector; the leave-one-dataset-out result provides the separate estimate for a transferable configuration.

\FloatBarrier

\section{Discussion}
\label{sec:discussion}
This section interprets the results and clarifies their empirical contribution. \Cref{subsec:discussion-contribution} distinguishes the study from a conventional standalone benchmark comparison. \Cref{subsec:discussion-complementarity,subsec:discussion-inference} discuss retrieval complementarity and paired inference; \cref{subsec:discussion-default-versus-domain,subsec:discussion-static-embeddings} contrast transferable and domain-specific fusion and assess static embeddings. \Cref{subsec:discussion-engineering} addresses engineering implications.

\subsection{Empirical contribution}
\label{subsec:discussion-contribution}
The contribution is not a new fusion algorithm: BM25--dense RRF is already an established retrieval pattern. The empirical contribution concerns marginal ranking value in a Dutch multilingual setting. The static retrievers are not merely weaker standalone models; under training-only selection they are rejected in every fold, and constraining the fusion to include them reduces held-out effectiveness. Public benchmark datasets make this result reproducible, but they do not predetermine the component rankings, selected weights, paired effects, or cross-domain transfer outcome.

\subsection{Within-dataset complementarity}
\label{subsec:discussion-complementarity}
BM25 and Qwen provide complementary ranking information on four datasets. The largest gain occurs on Dutch News, where equal weighting improves MRR by approximately 10.2\% relative to BM25. VABB and Wikipedia NL also benefit from equal weighting despite Qwen being the stronger standalone model. Open Tender remains lexical, while WebFAQ NL is predominantly semantic and receives only a small benefit from BM25.

\subsection{What the inferential results add}
\label{subsec:discussion-inference}
The paired tests do not replace effect sizes. They answer whether the observed out-of-fold query differences are compatible with a zero-effect null under the sign-randomisation design. Bootstrap CIs answer a different question by quantifying the plausible size of the average improvement. Their combination is especially useful for WebFAQ NL: the large query count makes a small effect detectable, while the CI makes clear that the expected gain remains operationally modest.

\subsection{A robust default versus a domain optimum}
\label{subsec:discussion-default-versus-domain}
The within-dataset selections show genuine domain dependence, but leave-one-dataset-out selection repeatedly chooses equal BM25--Qwen weighting. These results are not contradictory. A domain-specific optimum can outperform the robust default when representative in-domain queries are available, while $(0.5,0.5)$ remains a defensible configuration when the target domain is unknown or validation data are unavailable.

\subsection{Static embeddings}
\label{subsec:discussion-static-embeddings}
The static retrievers are weaker standalone systems and never receive positive weight from the unrestricted selector. Within the evaluated effectiveness-only setting, they therefore do not justify a third fusion component. This is a scoped architectural conclusion rather than a universal claim that static embeddings are unnecessary: other static models, retrieval-specific training, or deployment constraints may change the trade-off. In particular, static embeddings may remain valuable under latency, memory, energy, or processor-only constraints, but that case requires an explicit efficiency--effectiveness evaluation rather than an MRR argument.

\subsection{Engineering implications}
\label{subsec:discussion-engineering}
The pipeline demonstrates that exhaustive fusion experiments can be made tractable using memory-mapped arrays, batched matrix multiplication, compressed Parquet storage, and lazy DuckDB views. The approximately 103~GB score directory also makes the storage cost of exhaustive full-corpus scoring explicit; exact storage and runtime measurements are documented in Appendix~\ref{app:reproducibility}. Truncated retrieval would reduce this footprint but would introduce a candidate-depth approximation. Explicit identifier and row-alignment checks remain essential: a silent mismatch could invalidate dense scores while leaving aggregate outputs superficially plausible.

\section{Limitations}
\label{sec:limitations}
The experiments provide a controlled comparison of lexical, dense, static, and fused retrieval approaches across several Dutch retrieval benchmarks. Nevertheless, the conclusions should be interpreted within the scope of the evaluated datasets, relevance structure, retrieval configurations, and experimental setting. The principal limitations are as follows.

\begin{enumerate}
\item \textbf{Random query splits.} The 10-fold cross-validation estimates performance on unseen queries drawn from the same benchmark distributions. The evaluation does not explicitly model temporal, organisational, or topic-based distribution shifts that may occur in deployment.

\item \textbf{Evaluated domains.} Leave-one-dataset-out evaluation examines transfer across five Dutch retrieval datasets. Consequently, the resulting cross-domain conclusions are limited to the range of domains represented by these datasets.

\item \textbf{Inferential assumptions.} The paired bootstrap and sign-randomisation procedures treat queries as exchangeable observations. Substantial dependence between related or near-duplicate queries could affect the resulting uncertainty estimates.

\item \textbf{Single relevant document.} Each query has one relevant document, making MRR appropriate and numerically equivalent to MAP in this setting. The conclusions may not directly extend to tasks with multiple relevant documents, graded relevance, or recall-oriented evaluation objectives.

\item \textbf{RRF configuration search.} Weighted RRF is evaluated with $k=60$ and weight increments of 0.1. Other RRF constants or a finer weight grid are outside the scope of the present experiments.

\pagebreak 

\item \textbf{Exhaustive-ranking setting.} Fusion is evaluated over full-corpus rankings, avoiding candidate truncation as an experimental confounder. Production systems that fuse only top-$k$ candidate lists may exhibit somewhat different behaviour.

\item \textbf{Static-model and preprocessing scope.} Only two multilingual static models are evaluated, and Static Similarity is not retrieval-specialised. The results establish no complementary effectiveness for these models under the tested preprocessing choices, including Dutch stopword removal and stemming, Qwen prompting, title--body concatenation, and the available benchmark relevance judgements. They do not establish that all static representations must fail in every hybrid retrieval setting.

\item \textbf{Operational performance.} Encoding throughput, index size, retrieval latency, memory consumption, and energy use are not evaluated. The experiments therefore support conclusions about retrieval effectiveness rather than operational efficiency.
\end{enumerate}

These limitations primarily constrain the generalisability and operational interpretation of the findings rather than the internal comparison between the evaluated retrieval methods. Within the defined experimental setting, all methods are assessed using the same datasets, relevance judgements, ranking depth, evaluation metric, and cross-validation procedure.

\section{Recommended Extensions}
\label{sec:recommended-extensions}
The first extension should repeat the 10-fold experiment with several independent partitions. Where metadata permit, group-aware or temporal folds should replace random query splits to approximate the intended deployment boundary. Additional Dutch datasets would strengthen the leave-one-dataset-out conclusion and permit more credible domain-level inference.

A finer search around the selected BM25--Qwen edge, alternative RRF constants, and top-$k$ candidate fusion would test whether the current optimum is an artefact of the grid or exhaustive-ranking design. Additional effectiveness metrics become necessary when future datasets contain multiple relevant documents or graded labels; they are not required merely to duplicate the present single-rank target.

Having established that the evaluated static branch does not improve effectiveness, a direct follow-up can restrict fusion to BM25 and one transformer dense retriever, reducing the weight search to a single parameter. Comparing several embedder families and model sizes on the same Dutch tasks would show whether larger or stronger standalone encoders also provide greater complementary value after fusion. Reporting encoding throughput, retrieval latency, memory, index size, and energy alongside MRR would then produce an effectiveness--compute frontier and identify the smallest non-dominated transformer configuration.

\section{Conclusion}
\label{sec:conclusion}
This paper provides a controlled evaluation of whether modern multilingual static embeddings add retrieval effectiveness beyond BM25--transformer fusion across Dutch MTEB retrieval tasks. The contribution is empirical: weighted RRF is used to test component complementarity under training-only selection, paired query-level inference, and cross-dataset transfer.

\pagebreak

With respect to RQ1, CV-selected fusion outperforms the individual retriever selected on the same training queries on four of the five datasets. The held-out dataset improvements are 0.061 MRR on Dutch News, 0.029 on VABB, 0.004 on WebFAQ NL, and 0.025 on Wikipedia NL. Open Tender remains unchanged because BM25 alone is selected. All four positive differences remain distinguishable from zero after Holm correction, although the WebFAQ NL gain is small in practical terms.

Regarding RQ2, neither evaluated static embedding model provides measurable complementary effectiveness beyond BM25--Qwen fusion. The unrestricted selector assigns zero weight to the static component in all 50 fold-level selections, placing every selected configuration on the BM25--Qwen edge. Moreover, forcing a positive static contribution reduces effectiveness relative to that edge. Within the evaluated setting, the static models therefore add architectural and computational complexity without corresponding retrieval value.

With RQ3, the selected fusion weights are completely stable within the present experiment. Each dataset selects the same configuration in all ten folds, and the apparent full-data optimum coincides with the cross-validated estimate to the reported precision. This stability strengthens the empirical conclusions, but it remains a property of the evaluated datasets and should not be interpreted as a general argument against held-out model selection.

As for RQ4, leave-one-dataset-out selection chooses equal BM25--Qwen weighting in every iteration. This transferred configuration outperforms the individual retriever selected on the remaining four datasets for every held-out task. Equal weighting is therefore a robust tested default when the target domain is unknown, while dataset-specific tuning remains preferable when representative in-domain queries are available.

Finally, regarding RQ5, dataset-adaptive CV-selected fusion achieves the highest macro-average MRR, followed by Qwen, BM25, Static Similarity, and Potion. The main practical distinction is therefore not between fusion and one universally superior individual retriever, but between two deployment regimes: selecting the BM25--Qwen weight using held-out target-domain queries when such data are available, and using equal weighting when they are not.

Returning to the overarching problem statement, the strongest expected Dutch retrieval effectiveness among the evaluated systems is obtained through a two-component lexical--transformer architecture. BM25 and Qwen provide complementary ranking information across most evaluated datasets, whereas the tested static embeddings do not contribute additional held-out effectiveness. Equal BM25--Qwen weighting also transfers reliably across the evaluated domains, providing a defensible domain-agnostic default.

These conclusions are deliberately scoped to the evaluated models, datasets, preprocessing choices, and effectiveness objective. They justify removing the tested static branch from subsequent Dutch embedder comparisons, while leaving efficiency-constrained or retrieval-specialised static models as separate questions. A natural next study is therefore a one-parameter comparison of transformer embedder families and model sizes on an effectiveness--compute frontier.


\clearpage
\printbibliography[heading=bibintoc,title={References}]

\clearpage

\begin{appendices}
\section{Data and Code Availability}
\label[appendix]{sec:data-code-availability}
Source code, configuration files, notebook execution order, and aggregate result tables are available at \repositoryurl. The experiments are implemented in sequential notebooks covering setup, data preparation, exhaustive scoring, weighted-RRF generation, and CV result analysis. 
Intermediate embeddings, BM25 indexes, Parquet score files, and DuckDB views are generated locally. Benchmark datasets and model checkpoints are obtained through the MTEB and Hugging Face ecosystems subject to their respective licences.

\section{Complete Simplex Heatmaps}
\label[appendix]{app:simplex-heatmaps}

In every panel, $\alpha$ is the BM25 weight, $\beta$ is the Qwen weight, and $\gamma$ is the static-model weight. Each hexagon is a full-query descriptive configuration; the outlined hexagon marks the highest MRR in that panel.

\begin{figure}[H]
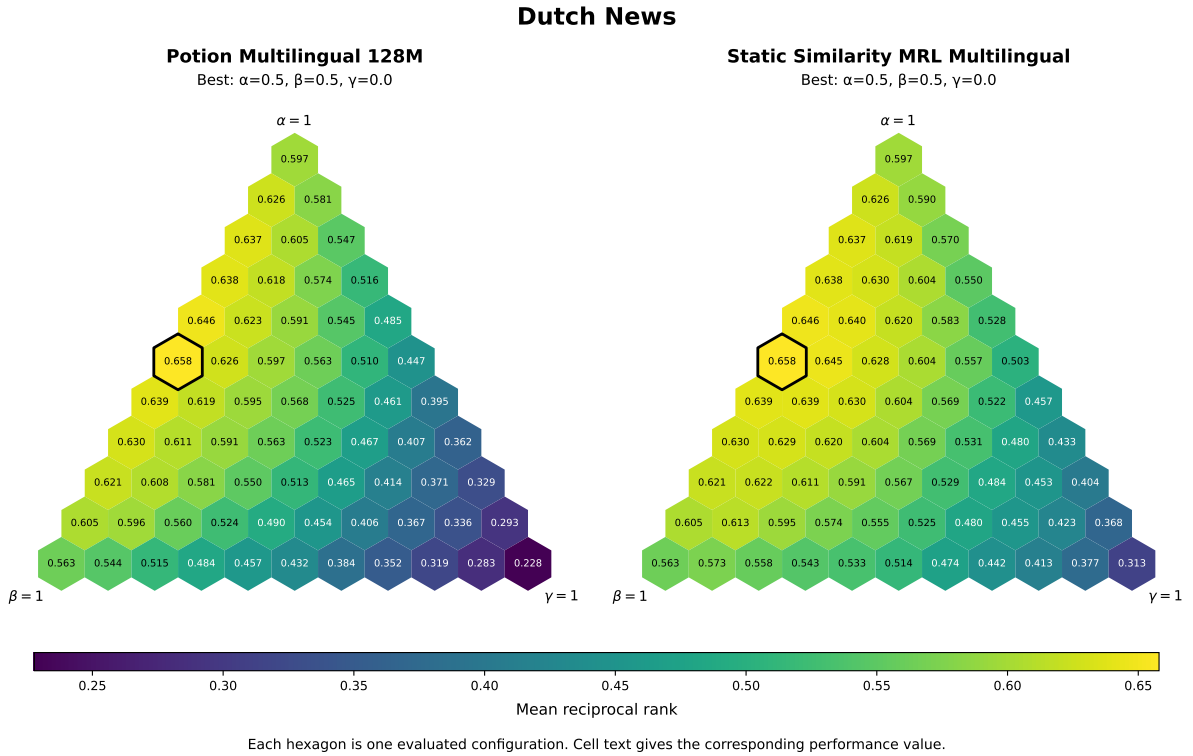

  \centering
  \resultgraphic{dutch_news_simplex.pdf}
  \caption{Dutch News descriptive simplex. Both panels are maximised at $(0.5,0.5,0.0)$ with MRR 0.658.}
\end{figure}

\begin{figure}[H]
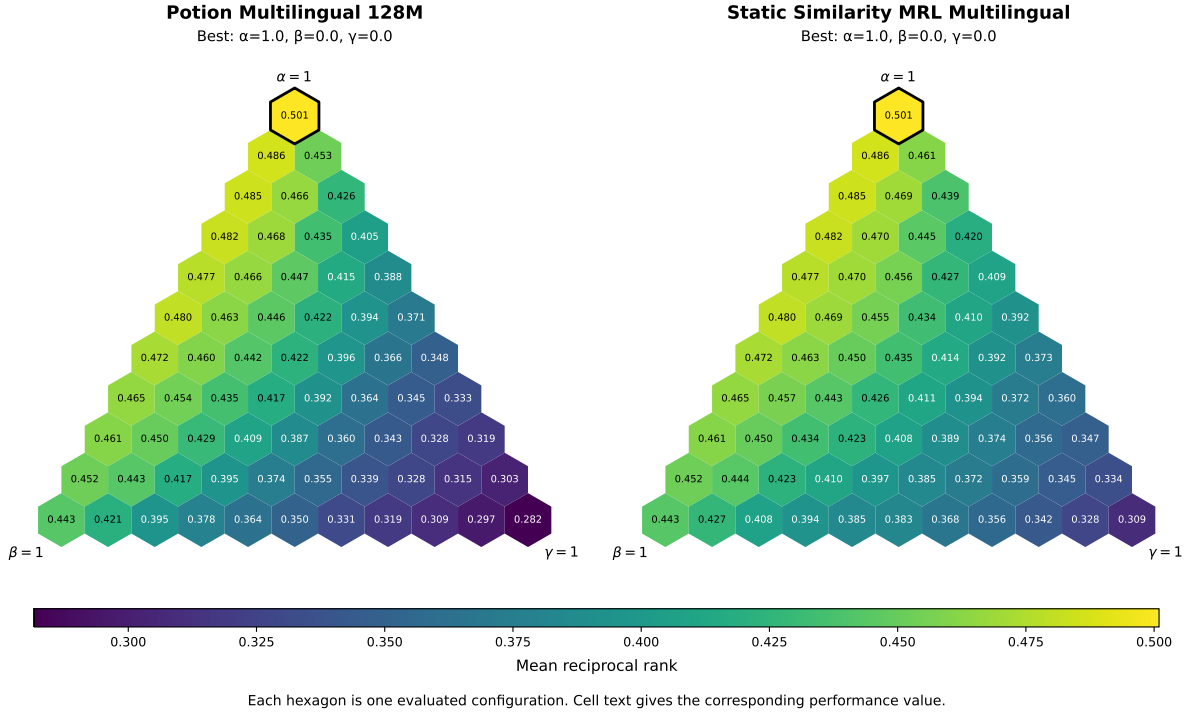

  \centering
  \resultgraphic{open_tender_simplex.pdf}
  \caption{Open Tender descriptive simplex. BM25 alone is optimal at $(1.0,0.0,0.0)$ with MRR 0.501.}
\end{figure}

\begin{figure}[H]
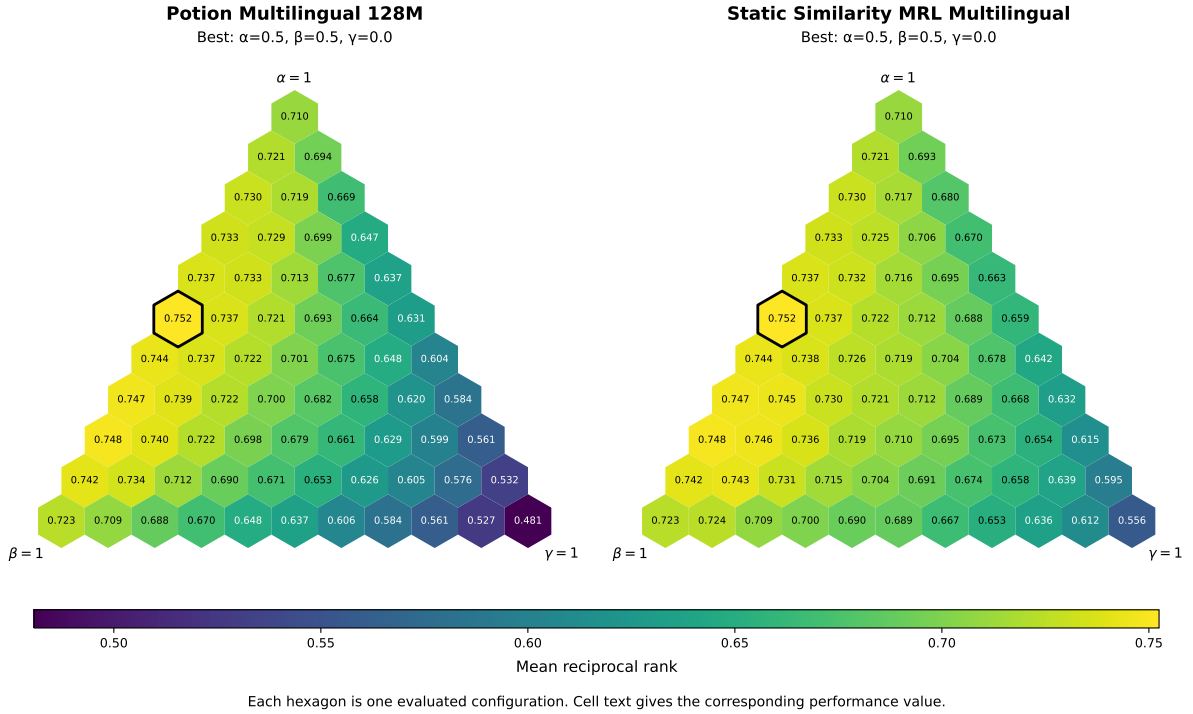

  \centering
  \resultgraphic{vabb_simplex.pdf}
  \caption{VABB descriptive simplex. Equal BM25--Qwen weighting is optimal at $(0.5,0.5,0.0)$ with MRR 0.752.}
\end{figure}

\begin{figure}[H]
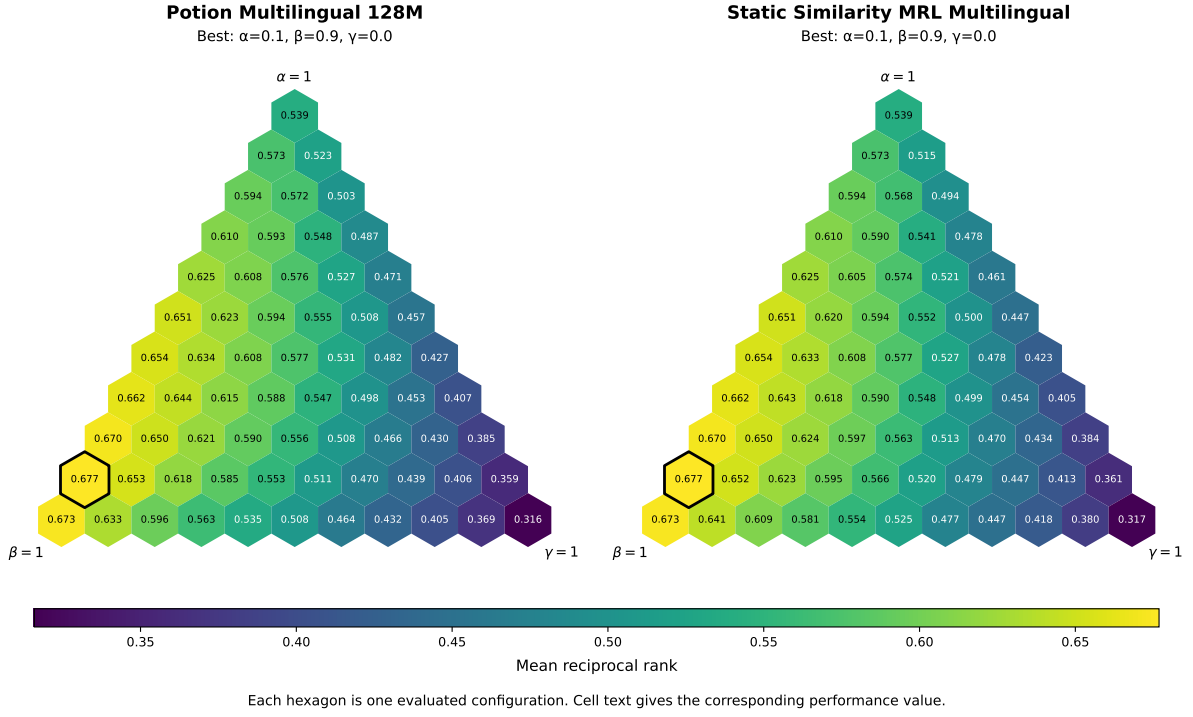

  \centering
  \resultgraphic{webfaq_nl_simplex.pdf}
  \caption{WebFAQ NL descriptive simplex. A Qwen-heavy mixture is optimal at $(0.1,0.9,0.0)$ with MRR 0.677.}
\end{figure}

\begin{figure}[H]
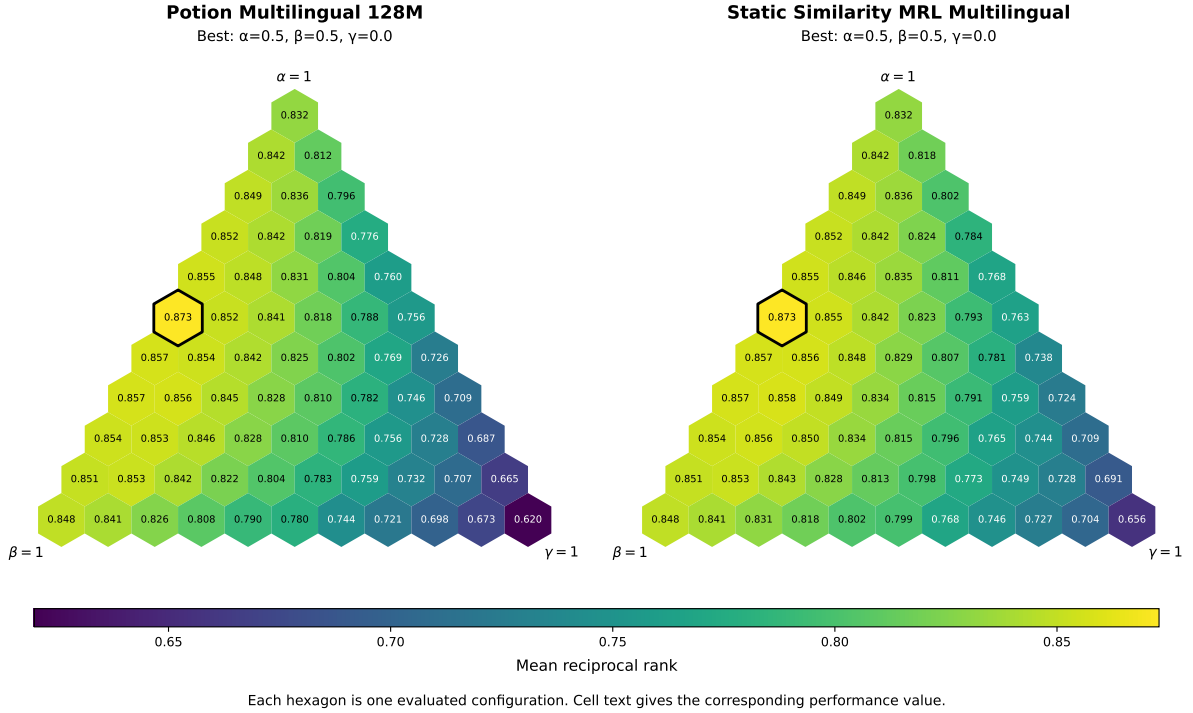

  \centering
  \resultgraphic{wikipedia_nl_simplex.pdf}
  \caption{Wikipedia NL descriptive simplex. Equal BM25--Qwen weighting is optimal at $(0.5,0.5,0.0)$ with MRR 0.873.}
\end{figure}

\FloatBarrier

\section{Reproducibility Details}
\label[appendix]{app:reproducibility}
This appendix records the resources needed to reproduce and interpret the experiment. Memory capacities are reported in gibibytes (GiB). \Cref{app:execution-environment,app:software-versions} document the execution environment and software versions; \cref{app:score-storage,app:runtime} report storage and measured runtime; and \cref{app:embedding-matrices} lists the stored embedding matrices.

\subsection{Recorded execution environment}
\label{app:execution-environment}
\begin{table}[H]
  \centering
  \caption{Recorded execution hardware and accelerator environment.}
  \begin{tabular}{@{}ll@{}}
    \toprule
    Component & Recorded value \\
    \midrule
    Operating system & Windows, amd64 \\
    Graphics processing unit (GPU) & NVIDIA RTX 4090 Laptop GPU \\
    GPU memory & 16.0 GiB \\
    PyTorch build & 2.11.0+cu128 \\
    Compute Unified Device Architecture (CUDA) runtime & 12.8 \\
    \bottomrule
  \end{tabular}
\end{table}

\subsection{Recorded software versions}
\label{app:software-versions}
\begin{table}[H]
  \centering
  \caption{Package versions recorded by the setup notebook.}
  \begin{tabular}{@{}lr@{}}
    \toprule
    Package & Version \\
    \midrule
    numpy & 2.5.1 \\
    pandas & 3.0.5 \\
    pyarrow & 25.0.0 \\
    duckdb & 1.5.5 \\
    mteb & 2.18.7 \\
    sentence-transformers & 5.6.1 \\
    model2vec & 0.8.2 \\
    bm25s & 0.3.10 \\
    matplotlib & 3.11.1 \\
    mpltern & 1.0.5 \\
    \bottomrule
  \end{tabular}
\end{table}

\subsection{Recorded score storage}
\label{app:score-storage}
\begin{table}[H]
  \centering
  \caption{Storage footprint of the exhaustive query--document score artifacts.}
  \label{tab:score-storage}
  \begin{tabular}{@{}lr@{}}
    \toprule
    Measure & Recorded value \\
    \midrule
    Logical file size & 111,487,959,025 bytes \\
    Size on disk & 111,488,000,000 bytes \\
    Windows display & 103 GB \\
    Binary-unit equivalent & approximately 103.83 GiB \\
    Decimal-unit equivalent & approximately 111.49 GB \\
    \bottomrule
  \end{tabular}
\end{table}

\subsection{Recorded wall-clock runtime}
\label{app:runtime}
The executed notebooks retain completed \texttt{tqdm} timers for exhaustive scoring and for rank fusion. These recorded stages sum to 3:30:56. The data-and-indexing notebook preserves only initial progress-widget states, while the final CV and statistical-analysis notebook contains no explicit elapsed-time measurement. The recorded total is therefore a lower bound on end-to-end execution and excludes dataset acquisition and preparation, embedding generation, BM25 index construction, and final result analysis.

\begin{table}[H]
  \centering
  \caption{Completed wall-clock timers retained by the scoring and fusion notebooks. Percentages refer to the 3:30:56 recorded total.}
  \label{tab:recorded-runtime}
  \small
  \begin{tabular}{@{}lrr@{}}
    \toprule
    Timed stage & Elapsed time & Share of recorded total \\
    \midrule
    Potion dense exhaustive scoring & 0:26:46 & 12.7\% \\
    Static Similarity dense exhaustive scoring & 0:27:43 & 13.1\% \\
    Qwen dense exhaustive scoring & 0:30:52 & 14.6\% \\
    BM25 exhaustive scoring & 0:29:28 & 14.0\% \\
    \addlinespace
    All exhaustive scoring & 1:54:49 & 54.4\% \\
    Fusion, ranking, and per-query metric generation & 1:36:07 & 45.6\% \\
    \midrule
    Combined recorded runtime & 3:30:56 & 100.0\% \\
    \bottomrule
  \end{tabular}
\end{table}

The nested fusion progress bars further account for 1:36:05 of the 1:36:07 fusion total. The remaining two seconds are outer-loop and bookkeeping overhead.

\begin{table}[H]
  \centering
  \caption{Dataset-level elapsed times retained within the fusion notebook.}
  \label{tab:fusion-runtime-by-dataset}
  \begin{tabular}{@{}lrr@{}}
    \toprule
    Dataset & Elapsed time & Share of fusion time \\
    \midrule
    Dutch News & 0:05:06 & 5.3\% \\
    Open Tender & 0:03:13 & 3.3\% \\
    VABB & 0:00:34 & 0.6\% \\
    WebFAQ NL & 1:25:47 & 89.2\% \\
    Wikipedia NL & 0:01:25 & 1.5\% \\
    Outer-loop overhead & 0:00:02 & less than 0.1\% \\
    \bottomrule
  \end{tabular}
\end{table}

\subsection{Stored embedding matrices}
\label{app:embedding-matrices}
Stored sizes are reported in mebibytes (MiB).

\begin{table}[H]
  \centering
  \caption{Float32 embedding matrix sizes recorded by the indexing notebook.}
  \small
  \begin{tabular}{@{}llrrr@{}}
    \toprule
    Model & Split & Rows & Dimension & Stored size \\
    \midrule
    Potion & Queries & 14,500 & 256 & 14.2 MiB \\
    Potion & Corpus & 786,573 & 256 & 768.1 MiB \\
    Static Similarity & Queries & 14,500 & 1024 & 56.6 MiB \\
    Static Similarity & Corpus & 786,573 & 1024 & 3,072.6 MiB \\
    Qwen & Queries & 14,500 & 1024 & 56.6 MiB \\
    Qwen & Corpus & 786,573 & 1024 & 3,072.6 MiB \\
    \bottomrule
  \end{tabular}
\end{table}

\end{appendices}

\end{document}